\documentclass[letterpaper, 10 pt, conference]{ieeeconf}  

\IEEEoverridecommandlockouts                              

\usepackage{epsfig} 
\usepackage{mathptmx} 
\usepackage{times} 
\usepackage{amsmath} 
\usepackage{amssymb}  
\usepackage{comment}
\usepackage{graphicx}
\usepackage{float} 
\usepackage{multirow}
\usepackage{algorithm}
\usepackage{algorithmic}
\usepackage{xcolor}
\usepackage{hyperref}
\hypersetup{pdfborder={0 0 0}}

\title{\LARGE \bf
A Comprehensive Framework for Automated Quality Control in the Automotive Industry
}

\author{Panagiota Moraiti$^{1}$, Panagiotis Giannikos$^{1}$, Athanasios Mastrogeorgiou$^{1,2}$, \\ Panagiotis Mavridis$^{1}$, Linghao Zhou$^{1}$, Panagiotis Chatzakos$^{3}$
\thanks{$^{1}$Panagiota Moraiti, Panagiotis Giannikos, Athanasios Mastrogeorgiou, Panagiotis Mavridis \& Linghao Zhou (\{panagiota.moraiti, panagiotis.giannikos, athanasios.mastrogeorgiou, panagiotis.mavridis, linghao.zhou\}@thlabs.eu) are with THL (Tech Hive Labs), 280 Kifisias Ave., 152 32 Halandri, GR.}%
\thanks{$^{2}$Athanasios Mastrogeorgiou (amast@central.ntua.gr) is with the Control Systems Lab, School of Mechanical Engineering, NTUA, GR.}%
\thanks{$^{3}$Panagiotis Chatzakos (p.chatzakos@essex.ai) is with the AI Innovation Centre, University of Essex, Wivenhoe Park, Colchester CO4 3SQ, UK.}
}

\begin{document}

\maketitle
\thispagestyle{empty}
\pagestyle{empty}

\begin{abstract}
This paper presents a cutting-edge robotic inspection solution (Fig. \ref{setup}) designed to automate quality control in automotive manufacturing. The system integrates a pair of collaborative robots, each equipped with a high-resolution camera-based vision system to accurately detect and localize surface and thread defects in aluminum high-pressure die casting (HPDC) automotive components. In addition, specialized lenses and optimized lighting configurations are employed to ensure consistent and high-quality image acquisition. The YOLO11n deep learning model is utilized, incorporating additional enhancements such as image slicing, ensemble learning, and bounding-box merging to significantly improve performance and minimize false detections. Furthermore, image processing techniques are applied to estimate the extent of the detected defects. Experimental results demonstrate real-time performance with high accuracy across a wide variety of defects, while minimizing false detections. The proposed solution is promising and highly scalable, providing the flexibility to adapt to various production environments and meet the evolving demands of the automotive industry.
\end{abstract}


\section{Introduction}
Quality control plays a crucial role in automotive manufacturing. Even minor defects introduced during production can result in significant performance issues and safety risks, emphasizing the importance of stringent quality inspections \cite{role_of_quality_control}. Traditionally, quality control processes in automotive production have been heavily dependent on skilled human operators to inspect components visually. This approach is not only costly and time-intensive but also susceptible to inconsistencies arising from operator fatigue and subjective decision-making \cite{automation_of_quality_control_DL}. In addition, manual inspection often struggles to meet the rising demands for precision and speed in modern manufacturing. Therefore, reliance on these traditional processes can hinder production efficiency and increase the risk of defective products reaching the market. To address these challenges, robotic solutions offer a promising alternative that delivers consistent, fast and accurate defect detection. By automating the inspection process, these advanced systems not only boost production efficiency, but also enhance the overall reliability of automotive manufacturing. 

\begin{figure}[t]
    \centering
    \includegraphics[width=0.49\textwidth]{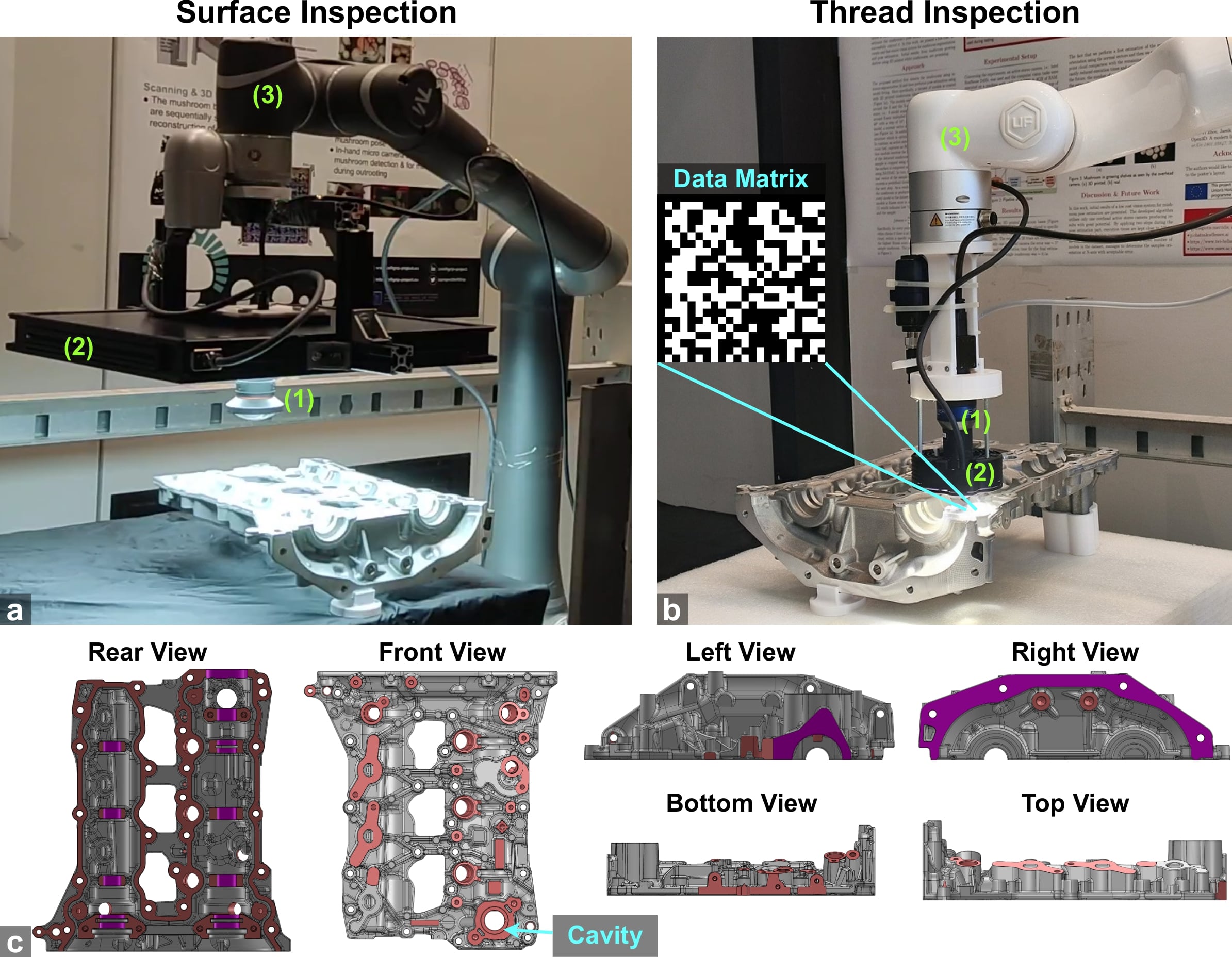}
    \caption{The developed quality control solution for inspecting surfaces and threads of aluminum HPDC automotive components. Each system consists of (1) a HD camera-based vision system, (2) a specialized lighting setup, and (3) a collaborative robot (cobot). a) The experimental setup fo surface inspection, and b) the experimental setup for thread inspection. c) The rear, front, left, right, and bottom views (five of the six sides) of the HPDC component. The front view features areas with wide cavity that are included in the inspection process.}
    \label{setup}
    \vspace{-3mm}
\end{figure}

The deployment of robotic systems in complex industrial environments poses several challenges, such as ensuring consistent image acquisition, optimizing speed, and achieving high detection accuracy across a wide range of defect types and sizes. Moreover, implementing artificial vision systems becomes difficult in cases of high geometric complexity of the parts.
Several strategies have been proposed for quality control in automotive manufacturing. One approach introduces a hierarchical system that leverages simulated data for automated inspection \cite{hierarchical_approach}. Another study \cite{metal_surface_defect_detection_ML} presents an advanced machine learning technique to detect and classify aluminum surface defects, a task commonly applied in various manufacturing contexts and closely related to the automotive industry. In publications such as \cite{automation_of_quality_control_DL, metal_surface_CNN, steel_metal_surface_DL} Convolutional Neural Networks (CNNs) are used for defect identification. Deep learning techniques for visual inspection in automotive assembly lines are explored in \cite{automotive_assembling_line}, which employs methods such as object detection, semantic segmentation and anomaly detection. A comprehensive review of state-of-the-art object detection models for steel defect detection is provided in \cite{steel_defects_detection_review}. Recent advancements include the “You Only Look Once" (YOLO) \cite{yolo} or “Single Shot MultiBox Detector" (SSD) \cite{ssd} architectures, utilized in works such as, \cite{wheel_surface_defect_detection, damaged_objects_car_body, defect_detection_automotive_ssd, GAN_based_car_body_yolov7}, and \cite{automotive_parts_yolov7}. Specifically, \cite{defect_detection_automotive_ssd} and \cite{automotive_parts_yolov7} enhance the SSD and YOLO algorithms, respectively, to optimize the detection of small defects. Additionally, \cite{lighting} proposes an illumination setup for optimal image acquisition, which is a crucial aspect of computer vision tasks that remains challenging due to the reflective nature of the components. However, all the mentioned approaches focus solely on computer vision and have yet to be integrated into a complete automated system for real-time quality inspection. 
Lastly, in \cite{mobile_robot}, a mobile robot uses a specific sensor to measure flush and gap in car bodies, but it does not address the highly demanding task of detecting defects in aluminum HPDC parts.

Our approach develops a comprehensive visual inspection framework and integrates it into a robotic solution. A scanning workflow is employed for image acquisition under optimized lighting configurations, designed to eliminate reflections on aluminum surfaces and ensure clear illumination for thread inspection. The overall solution for inspecting surfaces (Fig. \ref{setup}a) and threads (Fig. \ref{setup}b) of HPDC automotive parts comprises three main hardware components: (1) A high-resolution camera-based vision system, (2) A specialized lighting setup, (3) Two collaborative robots. For defect detection and localization, the YOLO11n \cite{yolo11} model is employed, offering high accuracy and robustness. Several enhancements have been integrated into the detection pipeline, including image slicing, ensemble learning, and bounding box merging, all aimed at improving detection performance and reliability. Additionally, a defect measurement module is incorporated to assess the severity of each detected defect, providing a comprehensive inspection framework. 

The main contributions of this work include: 1) a robust system that inspects both surfaces and threads of aluminum HPDC automotive components, 2) the integration of advanced techniques in the detection pipeline to improve accuracy and reduce false positives, and 3) the development of a deep learning based defect measurement module that estimates the size of the detected defects.

This paper is organized as follows: Sect. II describes the system setup, covering hardware components and software frameworks, while Sect. III outlines the automated inspection process, including the scanning, defect detection, and measurement methods. Sect. IV presents the experimental results, and Sect. V concludes with a discussion of future research directions.

\section{System Description}
This section provides an overview of the inspection system, detailing the hardware components and software frameworks utilized for automated defect detection.

\subsection{Hardware Components}
An area-scan camera with a CMOS global shutter sensor, operating at 60 fps, is selected to ensure accurate exposure and real-time image acquisition—crucial for high-speed industrial inspection tasks. Based on the smallest feature's size \(min\_feature\_mm = 0.5\) mm that is intended to be detected with \(min\_feature\_px = 10\) pixels in a $1/3$ of the automotive part's length meaning that \(FoV = 120\) mm, the desired resolution of the camera is:
\begin{equation}
    resolution = \frac{FoV \times min\_feature\_px}{min\_feature\_mm}
\end{equation}
that is slightly above the $2k$ resolution. As a result Hikrobot MV-CS050-10UC was chosen which has an essential $2448\times2048$ resolution, for capturing the desired detail. Two different types of lens are employed; the first is optimized to detect defects on large surfaces and the second is designed specifically to identify defects within holes. Taking into account: (1) the \(FoV = 120\) mm for surfaces, (2) the \(FoV = 30\) mm for threads, (3) the working distance between the camera and the object of interest \(WD = 86\) mm for surfaces, (4) the \(WD = 15\) mm for threads, (5) the \(resolution = 2448\) pixels and (6) the pixel size of the camera \(pixel\_size = 3.45~{\rm \mu m}\), the focal length for the lenses are:
\begin{equation}
    focal\_length = \frac{WD \times resolution \times pixel\_size}{FoV}
\end{equation}
Hence, the desired focal lengths are approximately $6$ mm for the surfaces and $4$ mm for the threads. In-depth market search resulted in opting for the Hikrobot MVL-KF0618M-12MPE and OptoEngineering PCHI023 lenses. The adjustable FoV, shutter, and focus offer great flexibility, enabling inspection of components of varying sizes and geometries.

To ensure clear and consistent imaging, while also eliminating reflections on aluminum HPDC components, the high-power white light flat dome HPFDOME $40$ cm $\times 40$ cm by TPLVision is utilized. The light accommodates a hole in its center for the installation of a camera and provides a white uniform $6500$ K color light under a white PMMA diffuser and with up to $45$ kLux in continuous operation. In addition, the selected lighting completely covers the required FoV and consequently it suits for the surface inspection of the component. Conversely, thread inspection necessitates compact illumination to avoid obstructing the motion of the vision system around the automotive component since for thread inspection the robot positions the lens at a distance of $15$ mm (compared to $86$ mm for surface inspection). To achieve this, a $37$ mm modular ring light (M-TRING) equipped with a C22 convergent angle changer from TPLVision was employed, effectively illuminating the inner cavities of the threads and producing high-quality imaging data.

A pair of collaborative robots (cobots) are utilized for both kinds of inspections (surface and thread). Specifically, a Techman TM12 6-DoF robotic arm is employed for surface inspection (Fig. \ref{setup}a), while a UFACTORY xArm 6-DoF robotic arm is used for thread inspection (Fig. \ref{setup}b).

\subsection{Software Frameworks}
The computer vision system utilizes YOLO11n \cite{yolo11} for defect detection and the Slicing Aided Hyper Inference (SAHI) library \cite{sahi} to enhance detection accuracy, particularly for small or subtle defects. The cobot operation is managed using NVIDIA cuRobo \cite{cuRobo} and ROS2 \cite{ROS2}, which enable efficient motion planning and real-time control. ROS2 facilitates seamless communication between system components, while cuRobo leverages GPU acceleration to optimize motion execution and ensure precise camera positioning during inspection.

\section{Automated Inspection}
This section outlines the automated inspection process, detailing the scanning procedure, defect detection and measurement methods, and the overall inspection workflow.

\subsection{Scanning Process}
As shown in Fig. \ref{setup}c, only five of the six sides of the automotive part—rear, front, left, right, and bottom—require inspection. Table \ref{number_of_images} summarizes the number of images needed for full coverage of each view, with $23$ images allocated for surface inspection and $198$ images for thread inspection, resulting in a total of $221$ images. The front view includes a wide cavity (Fig. \ref{setup}c), necessitating a cross-shaped trajectory for the camera to effectively inspect the inner walls. Specifically it takes $\sim 13.6$ s to scan the most complex surface of the part (Front View in Fig. \ref{setup}c) and $\sim 34.75$ s to acquire all the needed images from all the surfaces of the part. For thread inspection, an image is captured from above and at the center of each thread, while the cross-shaped trajectory is also employed, acquiring four additional images to ensure a thorough examination. The time required to acquire the images of one thread is $\sim5$ s, resulting in a total of $\sim 2.9$ min for the inspection of the $35$ threads of the part. The presence of a large amount of threads in the part makes thread inspection the most time-consuming process of the pipeline.

\begin{table}[ht]
\caption{Number of Images for Surface and Thread Inspection}

\label{number_of_images}
\begin{center}
\begin{tabular}{c c c c c}
\hline
View & \begin{tabular}[c]{@{}c@{}}Surface Images\end{tabular} & \begin{tabular}[c]{@{}c@{}}No of Threads\end{tabular} & \begin{tabular}[c]{@{}c@{}}Thread Images\end{tabular} & \begin{tabular}[c]{@{}c@{}}Total\end{tabular} \\
\hline
Rear & 7 & 7 & 35 & 42 \\
Front & 9 & 21 & 105 & 114 \\
Left & 2 & - & - & 2 \\
Right & 3 & 4 & 20 & 23 \\
Bottom & 2 & 3 & 15 & 17 \\
\hline
\end{tabular}
\end{center}
\end{table}


\subsection{Defect Detection}

The nano version of YOLO11 is well-suited to our requirements, as it provides an optimal balance between detection accuracy and inference speed, which is crucial for industrial applications. Defects are categorized into surface and thread types: scratches, dents, and irregularities on the outer layer belong to the surface category, while deformations or misalignment correspond to thread defects. To accurately detect both defect types, two specialized models are employed, each tailored to capture the unique features of its respective defect class.

\begin{figure}[ht]
  \centering
  \includegraphics[width=0.49\textwidth]{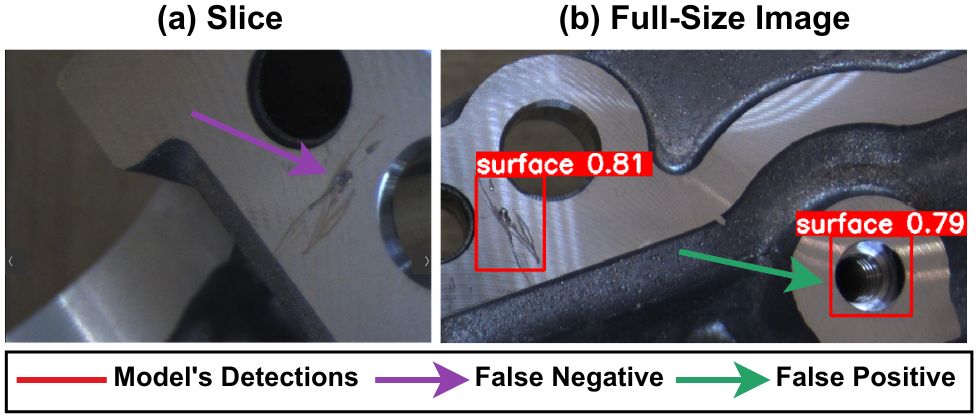}
  \caption{Detections from two models. (a) Models trained on full-size images excel at detecting small, subtle defects, but struggle to detect large defects in slices. (b) Models trained solely on slices may misclassify holes, as defects in full-size images.}
  \label{example}
\end{figure}

One challenge encountered was the model’s difficulty in detecting small or subtle surface defects, often failing to recognize them due to low confidence scores. To mitigate this issue, the Slicing Aided Hyper Inference (SAHI) framework \cite{sahi} was adopted with modifications to better align with the defect detection task. The original SAHI technique segments an image into smaller overlapping regions, upscales them, and applies detection to each segment, utilizing either the original model trained on full-size images or a fine-tuned model trained on a subset of slices extracted from the training dataset. Additionally, inference can be performed on the full-size image to detect larger objects that may not fit within a single slice. Although SAHI is effective for conventional object detection tasks, defect detection presents unique challenges due to significant scale variations and distinct feature distributions between full-size images and slices, as highlighted in \cite{small_defects_IC}. Specifically, defects in full-size images tend to be less distinguishable, whereas in sliced images, they appear larger and more prominent. Consequently, models trained on full-size images may fail to detect larger defects in slices (Fig. \ref{example}a), while models trained on slices may misinterpret regular features as defects in full-size images (Fig. \ref{example}b).

To address the aforementioned issue, the model was trained exclusively on slices extracted from full-size images. While missed defects were reduced, false positives increased  (Fig. \ref{FP_FN}, SAHI-V1), primarily due to black spots or stains, caused by manufacturing residues or marks introduced during the milling process. To mitigate these problems, we incorporated non-defective images containing these irregularities into the training set, which helped reduce false positives. However, this approach introduced more false negatives than before (Fig. \ref{FP_FN}, SAHI-V2). Training a single model with non-defective images alone does not fully resolve the issue. 

Alternatively, we propose training two separate models: one using non-defective images from the first category of false positives (Fig. \ref{FP_FN}, SAHI-V3), and the other from the second category (Fig. \ref{FP_FN}, SAHI-V4). The defective training images remain the same. This approach prevents the models from being overwhelmed by too many non-defective instances during training, allowing them to maintain high recall. Since they focus on different types of errors, their outputs differ significantly. Therefore, rather than relying on a single model, we combine predictions from both using ensemble learning (Fig. \ref{FP_FN}, Ensemble). 

In surface inspection, images are sliced into $4$ tiles of $1280\times1071$ pixels, a configuration determined through extensive experimentation. A higher number of slices results in excessive zooming, increasing the likelihood of small abnormalities being incorrectly detected as defects. The original image is $X \in \mathbb{R}^{H \times W \times 3}$ and each slice is $X_{\text{slice}}^{(i)} \in \mathbb{R}^{h \times w \times 3}$, where $H$ and $W$ are the height and width of the original image, and $h$ and $w$ are the dimensions of each individual slice. A forward pass is performed on these slices and an ensemble learning technique is utilized. The detections are first filtered using a confidence threshold of $0.7$ to retain only the most reliable ones. Next, only common detections from SAHI-V3 and SAHI-V4 with at least $1\%$ bounding box overlap, based on IoU (Intersection over Union), are retained. IoU is a metric that measures the overlap between two bounding boxes by dividing the area of their intersection by the area of their union. The detections are then merged using the traditional Non-Maximum Suppression (NMS) technique \cite{nms} with a $0.15$ threshold to eliminate redundant detections, preserving the bounding box with the highest confidence.

In thread inspection, only one model is utilized, and a forward pass is performed. Confidence filtering and NMS are then applied to refine the results, with thresholds of $0.65$ and $0.15$, respectively.

A comprehensive description of the defect detection process is outlined in Algorithm \ref{alg1}.

\begin{algorithm}[ht]
\caption{Defect Detection}
\label{alg1}
\begin{algorithmic}[1]
\STATE \textbf{Input:} Original image \( X \in \mathbb{R}^{H \times W \times 3} \), Models \( F_1 \), \( F_2 \), \( F_3 \), Conf Thresh \( C_{T_1}, C_{T_2} \), IoU Thresh \( S_T \), NMS Thresh \( N_T \)
\IF{Surface}
    \STATE Slicing: \( X \to \{ X_{\text{slice}}^{(i)} \in \mathbb{R}^{h \times w \times 3} \} \)
    \STATE \( P_1, P_2 \gets \emptyset \)
    \FOR{each slice}
        \STATE Forward Pass: \(\text{p}_1^{(i)} =  F_1(X_{\text{slice}}^{(i)}) \), \(\text{p}_2^{(i)} =  F_2(X_{\text{slice}}^{(i)}) \)
        \STATE Thresh: \( p_k^{(i)} \gets \{ p_k^{(i)} \mid \text{conf}(p_{k}^{(i)}) \geq C_{T_1} \}, \quad k \in \{1,2\} \)
        \STATE \( P_1 \gets P_1 \cup p_1^{(i)}, \quad P_2 \gets P_2 \cup p_2^{(i)} \)
    \ENDFOR
    \STATE Common Preds: \( \text{P} = \{ (P_{1,i}, P_{2,j}) \mid \text{IoU}(P_{1,i}, P_{2,j}) \geq S_T \} \)
\ELSIF{Thread}
    \STATE Forward Pass: \(\text{P} =  F_3(X) \)
    \STATE Thresholding: \( \text{P} \gets \{ \text{P} \mid \text{conf(P)} \geq C_{T_2} \} \)
\ENDIF
\STATE \( \text{P}_{\text{final}} = \text{NMS}(\text{P}, N_T) \)
\STATE \textbf{Output:} \( P_{\text{final}} \)
\end{algorithmic}
\end{algorithm}

\subsection{Defect Size Measurement}\label{section:3c}
The inspection procedure continues with measuring the size of each detected defect. Quality standards refer to the size of a defect as equal to its diameter, which in pixels can be approximated as:
\begin{equation}
    defect\_px = \sqrt{\sqrt{w^2 + h^2} \times max(w, h)}
    \label{eq:defect size in pixels}
\end{equation}
where $w$ and $h$ are the width and height dimensions of the bounding box in pixels, provided by the detection algorithm and $max$ refers to the maximum function.

The second step in determining the size of a defect in an image is to perform a calibration using a reference object; not to be confused with the intrinsic or extrinsic calibration of the camera. The reference object should have known dimensions in a quantifiable real-world unit and should be easily located in an image, either by its placement or by its appearance \cite{dimensional_analysis}. Along these lines, our choice was a $10$ mm $\times 10$ mm data matrix, shown in Fig. \ref{setup}, which is not the real one, but one generated for the purposes of this publication. The reference dimension is \(reference\_mm = 10\) mm, which additionally encodes information about the specific type of the automotive component. At the beginning of the inspection procedure, the robotic system locates the data matrix of the component (see Fig. \ref{setup}b), calculates its dimensions in pixels \(reference\_px = 160\) pixels and \(reference\_px = 958\) pixels for surface and thread inspection respectively, and determines the millimeters per pixel ratio \(reference\_mm/reference\_px = 0.0623\) mm/pixel and \(reference\_mm/reference\_px = 0.0104\) mm/pixel for surface and thread inspection respectively. The ratio is key to translating every distance in pixels within the image to the respective distance in millimeters. This enables us to convert the approximation of a defect's diameter from pixels to millimeters:
\begin{equation}
    defect\_mm = \frac{defect\_px \times reference\_mm}{reference\_px}
    \label{eq:defect size in mm}
\end{equation}

\subsection{Inspection Workflow}
\begin{figure}[ht]
  \centering
  \includegraphics[width=0.48\textwidth]{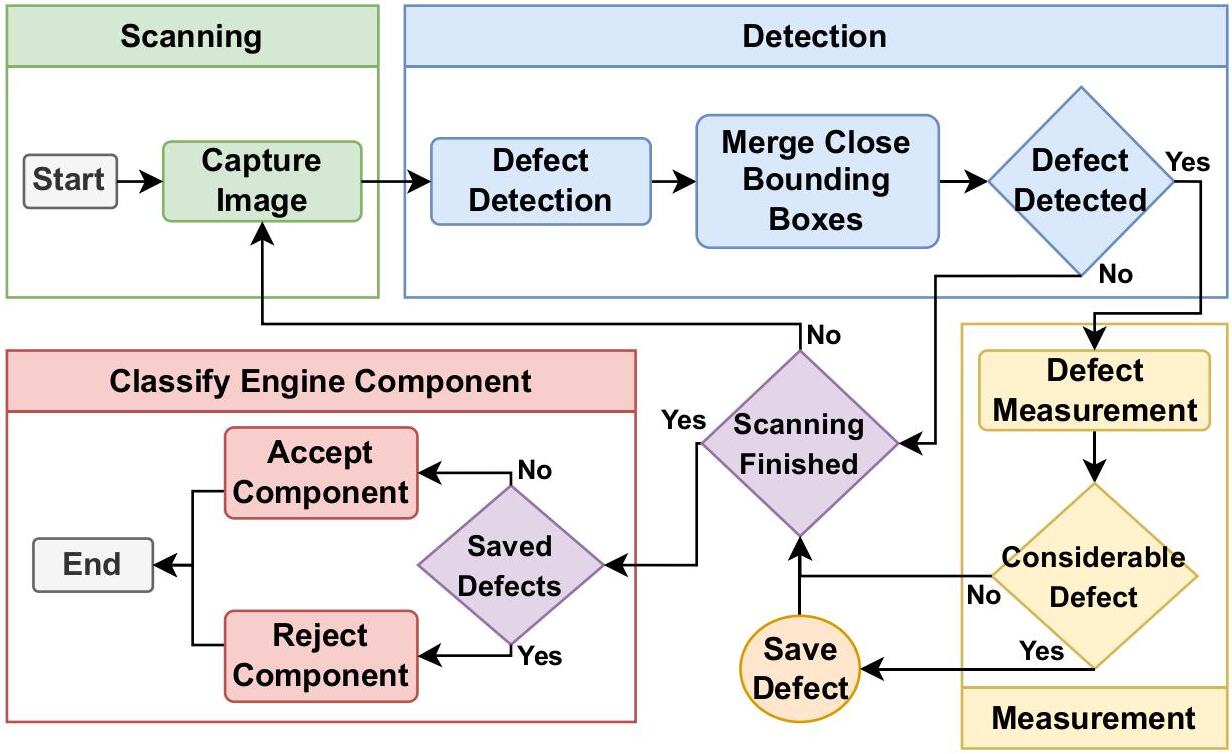}
  \caption{The inspection workflow begins with the scanning of the aluminum HPDC component, followed by defect detection and measurement to assess the size and severity of any detected defect. The final step is the classification module, where the part is classified as defective if any considerable defect is found.}
  \label{workflow}
\end{figure}

The proposed inspection workflow consists of four stages, as illustrated in Fig. \ref{workflow}. The process begins with the scanning of the aluminum HPDC component, followed by defect detection using our ensemble model for the surfaces and YOLO11n for the threads. Then, the bounding box merging module combines close detections to cover the entire region. This step is necessary because multiple defects gathered in a small area may be too small individually, but collectively they could represent a considerable issue. The proximity of the detections is determined using the Euclidean distance between their centers:
$d = \sqrt{\sum_{i=1}^{2} (x_i - y_i)^2}$, where \((x_1, y_1)\) and \((x_2, y_2)\) are the centers of the two bounding boxes. If \( d \leq 20 \) pixels for surface inspection, which corresponds to $1.246$ mm, the bounding boxes are merged. For thread inspection, the threshold is set to $120$ pixels, corresponding to $1.248$ mm. The resulting bounding box spans the full extent of the clustered defects, retaining the highest confidence score among them.

Next, the measurement module quantifies the size of each detected defect. The maximum accepted size is defined a priori, depending on the use-case requirements. For test purposes, we set it as $2$ mm. If a defect's size exceeds this threshold, the defect is flagged as considerable. Once the scanning is completed, if a considerable defect has been found, the part is rejected. The size and location of the defects on the component, as well as the overall classification of the part, are recorded in a CSV file for later quality control analysis.

\section{Experimental Results}
This section presents the datasets and evaluation metrics utilized, followed by a comprehensive analysis of the experiments conducted for surface and thread defect detection and measurement. Table \ref{results} and Fig. \ref{comparison} show the results of the defect detection procedure. 

\subsection{Dataset}
To train the defect detection models, we collected and annotated a dataset of defects from $8$ aluminum HPDC components. In order to improve the model’s robustness to small variations, images are captured under varying lighting conditions and camera exposure times. The dataset includes $158$ surface images and $94$ thread images. Both datasets are split into training, validation, and test sets. The test sets contain defect instances not seen during training or validation, ensuring the model generalizes to new, unseen defects.


To better evaluate false positives, $116$ non-defective surface images and $45$ thread images were added to the test sets. False positives are critical, as even a single defect detection can lead to rejecting the entire component, resulting in unnecessary costs.

Data augmentation was applied to improve the model's generalization and prevent overfitting. The augmentations included random brightness adjustments, horizontal and vertical translations, scaling to accommodate varying object sizes, and flipping to help generalize across different orientations. We experimented and found that, an input size of $1280\times1280$ pixels provided the best results for the current inspection task.



\subsection{Evaluation Metrics}
Average Precision (AP) is a standard metric in object detection, computed as the area under the Precision-Recall (P-R) curve. AP reflects how well the model balances precision (accurate detections) and recall (correctly detecting all relevant objects). The mean Average Precision (mAP) is the average of the AP scores across all object classes.
The evaluation of the models is based on the mAP50 and mAP30 metrics, which measure the mAP at IoU thresholds of $50\%$ and $30\%$, respectively. This means that mAP50 requires at least a $50\%$ overlap (\( \text{IoU} \geq 0.5 \)) between predicted and ground truth bounding boxes to be considered a correct detection, while mAP30 allows for a lower threshold (\( \text{IoU} \geq 0.3 \)), useful for detecting subtle or ambiguous defects.


\subsection{Surface Defect Detection}
\begin{figure}[ht]
  \centering
  \includegraphics[width=0.48\textwidth]{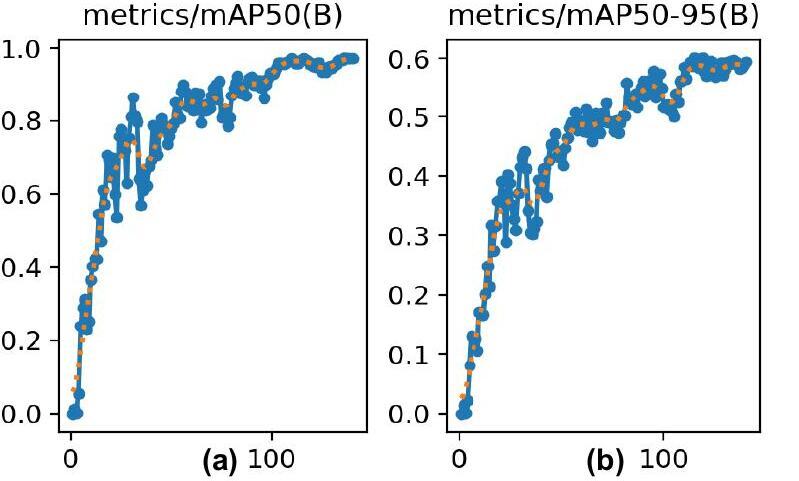}
  \caption{(a) mAP50 and (b) mAP50-95 curves for YOLO11n in surface defect detection. After approximately epoch $125$, the metrics start to converge, indicating that further training does not yield significant improvements in model's performance.}
  \label{map}
\end{figure}

For surface defect detection, the YOLO11n pre-trained model was utilized. Following the official guidelines for effective transfer learning, which recommend a minimum of $100$ annotated images and approximately $100$ epochs for training in the case of a single class \cite{ultralytics_guide}, we trained the model for $140$ epochs, using early stopping to prevent overfitting. The mAP curves for YOLO11n are shown in Fig. \ref{map}. After epoch $125$, the metrics begin to converge, indicating no significant improvement with additional training. Despite achieving high accuracy, some false positives and false negatives remain. mAP30 is $94.1\%$ and mAP50 is $55.1\%$, with a total of $9$ incorrect detections, as shown in Table \ref{results}.

\begin{figure*}[ht]
    \centering
    \includegraphics[width=1\textwidth]{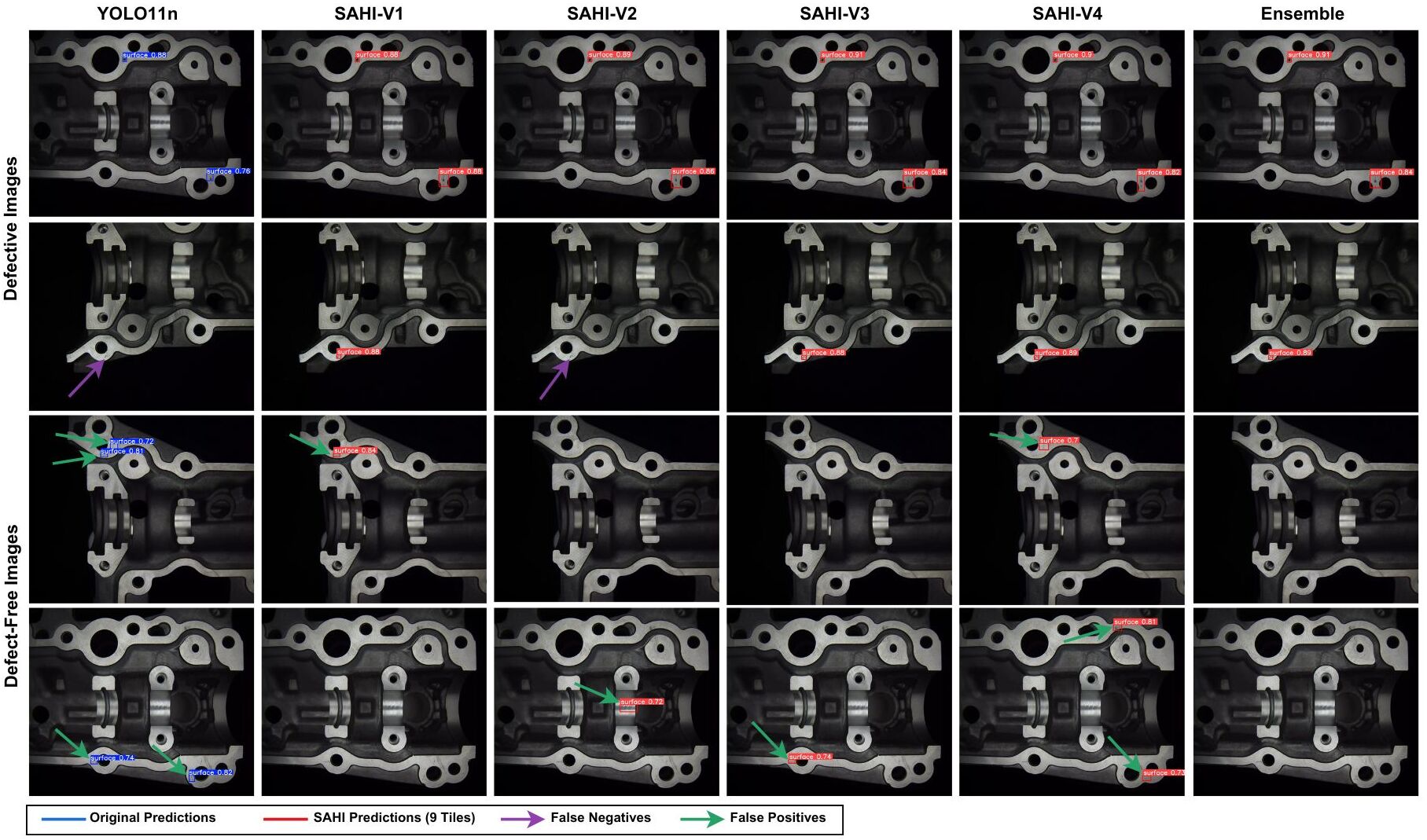} 
    \caption{Comparison of surface defect detection across different training settings. The captured images are of high quality, with no reflections, proper lighting, and clear visibility of the defects. In SAHI-V1, the slicing technique is utilized, and only defective images are used during training. In SAHI-V2, non-defective images are incorporated into the training process. In SAHI-V3 and SAHI-V4, non-defective images of black stains and internal marks are utilized during training, respectively. Finally in Ensemble, only the common detections of SAHI-V3 and SAHI-V4, which are trained on different subsets of non-defective images, are retained. The Ensemble model successfully detects all defects correctly in these examples.}
    \label{comparison}
\end{figure*}

To further enhance performance, the SAHI framework was utilized. Training and validation images were divided into $4$ tiles, and only those containing at least one defect were retained to focus training on relevant defect patterns. During inference, the image is divided into $9$ tiles with an overlap ratio of $0.25$. This ensures that defects are fully captured, preventing them from being missed or fragmented. During training, no overlapping were used, allowing fragmented defects to serve as challenging instances that contribute to the robustness of the model. After tiling, the training and validation set consist of $193$ images, which are resized to $1280\times1280$ pixels using padding. Now, defects become more prominent, making the detection task easier compared to the original images. Training for more epochs led to overfitting, which resulted in a significant number of missed defects in the test images. As a result of this approach, the SAHI-V1 model demonstrates strong classification accuracy and enhanced localization capabilities. The total number of false positives is $16$, reflecting a decrease in precision. However, there are no false negatives, confirming that the model successfully identifies all the defects, indicating a high recall. The trade-off between precision and recall is common in detection tasks, where prioritizing high recall can sometimes lead to a decrease in precision.

To reduce false positives, $19$ non-defective tiles are added to the training and $8$ to the validation set. Training on these images teaches the SAHI-V2 model to ignore abnormalities, such as black stains and internal marks within the aluminum, that may resemble defects and results in a total of $10$ incorrect detections. Although this model minimizes the total number of false positives, this comes at the cost of introducing some false negatives. Therefore, this approach does not yield better results. 

After careful analysis of the detections and identification of weaknesses in the dataset, two additional models were trained, and the false positives were categorized into two distinct types. Each model was tailored to specialize in ignoring only one category, improving overall detection reliability. The key idea is to retain only the common predictions of the two models, minimizing false detections. For the overall proposed method to be effective, both models must achieve very high recall, ensuring that all defects are detected. For the first model, $7$ non-defective tiles from the first category were added to the training set and $3$ to the validation set, while for the second model, $12$ non-defective tiles from the second category were added to the training set and $5$ to the validation set. 
These models, SAHI-V3 and SAHI-V4, demonstrate high recall with no missed defects but still produce several false positives. Extensive analysis revealed that each model has learned to ignore different types of abnormalities. As a result, the predictions from both are combined using the ensemble learning technique mentioned above.
The final model, referred to as Ensemble, achieves an mAP30 of $99.7\%$ and an mAP50 of $78.4\%$, demonstrating improved performance by leveraging the strengths of both models (SAHI-V3 and SAHI-V4). While the localization capabilities, as indicated by the mAP50, may not be optimal, our main concern for this task is the classification of the components as defective or defect-free. Overall, our proposed method outperforms the original YOLO11n in both classification and localization tasks. Our approach can be utilized in visual inspection where false positive detections pose challenges and is particularly effective even in data-limited scenarios. Table \ref{results} shows the comprehensive results of each method utilized for surface defect detection.

\begin{figure}[ht]
  \centering
  \includegraphics[width=0.48\textwidth]{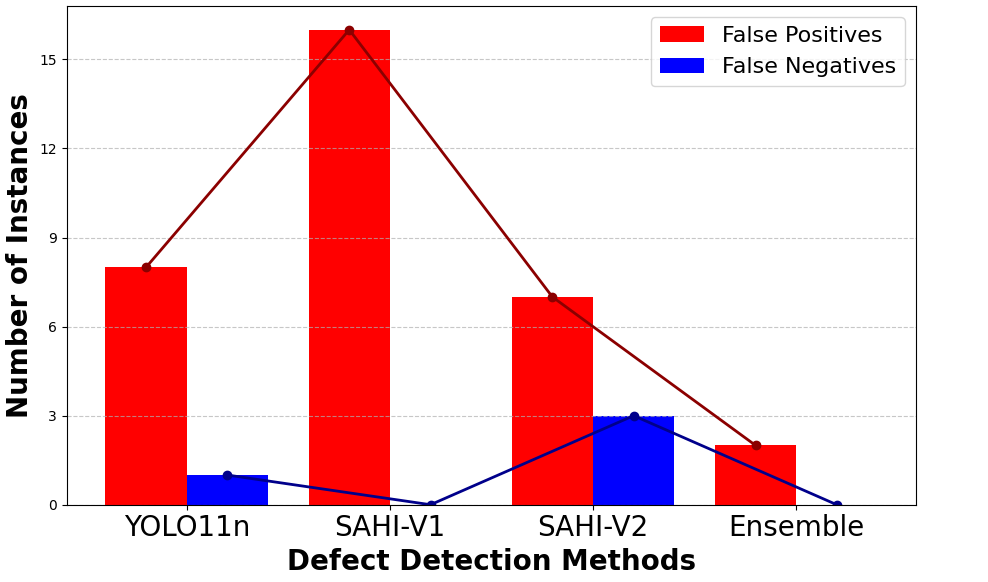}
  \caption{Total number of False Positives (FP) and False Negatives (FN) observed in surface defect detection across different training settings, with an IoU threshold of $0.3$. YOLO11n detects 8 FP and 1 FN, while SAHI-V1 minimizes the FN, but the FP are increased. SAHI-V2 does not yield better results compared to YOLO11n. The ensemble model retains only common detections from SAHI-V3 and SAHI-V4, resulting in zero FN and two FP, which is the best performance achieved.}
  \label{FP_FN}
\end{figure}

In Fig. \ref{comparison}, examples of surface defect detections are shown and in Fig. \ref{FP_FN}, a graph is presented that visualizes the total number of false positives and false negatives observed in surface defect detections for different training settings, with an IoU threshold of $0.3$. 

Overall, the predictions of YOLO11n include $8$ false positives and $1$ false negative occurs, while SAHI-V1 minimizes missed defects to $0$ but detects $16$ false positives. SAHI-V2 does not yield better results with $7$ false positives and $3$ false negatives. SAHI-V3 and SAHI-V4, trained on different non-defective image subsets, eliminate missed defects but introduce several false positives ($8$ and $19$, respectively). Finally, the ensemble model retains only common detections from SAHI-V3 and SAHI-V4, ensuring all defects are detected and minimizing false positives to $2$, which is the lowest number achieved.

\subsection{Thread Defect Detection}
The YOLO11n pre-trained model was trained for $93$ epochs with early stopping to prevent overfitting. Training was terminated when the validation metrics began to converge. 
The model achieves an mAP30 of $89.1\%$ and an mAP50 of $83.8\%$, with $2$ missed defects, as shown in Table \ref{results}. Since the images are captured from a close range and no false positives occur, neither the SAHI technique nor our proposed ensemble method is necessary for this task. 

A small number of false negatives is acceptable, because the inspection involves capturing five views of the same thread, as mentioned above. Therefore, the defect must be detected in at least one of these. In our test set, the $2$ false negatives correspond to challenging views of an instance already detected in a different image, meaning the overall recall of the system is actually higher.

\renewcommand{\arraystretch}{1.2} 
\begin{table}[ht]
    \setlength{\tabcolsep}{4.5pt}
    \caption{Results for Surface and Thread Defect Detection}
    
    \label{results}
    \begin{center}
    \begin{tabular}{c c c c c c c c c}
    \hline
    & \multicolumn{1}{c}{\multirow{2}{*}{Method}} & \multicolumn{1}{c}{Training} & \multicolumn{1}{c}{Validation} & \multicolumn{4}{c}{Test} \\
    \cline{3-8}
    & & mAP50 & mAP50 & mAP50 & mAP30 & FP & FN \\
    \hline
    \multirow{5}{*}{\textbf{\rotatebox{90}{Surface}}} & YOLO11n & 0.832 & 0.651 & 0.551 & 0.941 & 8 & 1 \\
    & SAHI-V1 & 0.98 & 0.941 & 0.937 & 0.983 & 16 & 0 \\
    & SAHI-V2 & 0.97 & 0.941 & 0.59 & 0.852 & 7 & 3 \\
    & SAHI-V3 & 0.967 & 0.941 & 0.744 & 0.995 & 8 & 0 \\
    & SAHI-V4 & 0.948 & 0.941 & 0.811 & 0.91 & 19 & 0 \\
    & Ensemble & - & - & 0.784 & \textbf{0.997} & \textbf{2} & \textbf{0} \\
    \hline
    \hline
    & \textbf{Thread} & \multirow{2}{*}{0.929} & \multirow{2}{*}{0.856} & \multirow{2}{*}{\textbf{0.838}} & \multirow{2}{*}{\textbf{0.891}} & \multirow{2}{*}{0} & \multirow{2}{*}{2} \\
    & YOLO11n &  &  &  &  &  &  \\
    \hline
    \end{tabular}
    \end{center}
\end{table}

\subsection{Defect Size Measurement}

\begin{figure}[ht]
  \centering
  \includegraphics[width=0.48\textwidth]{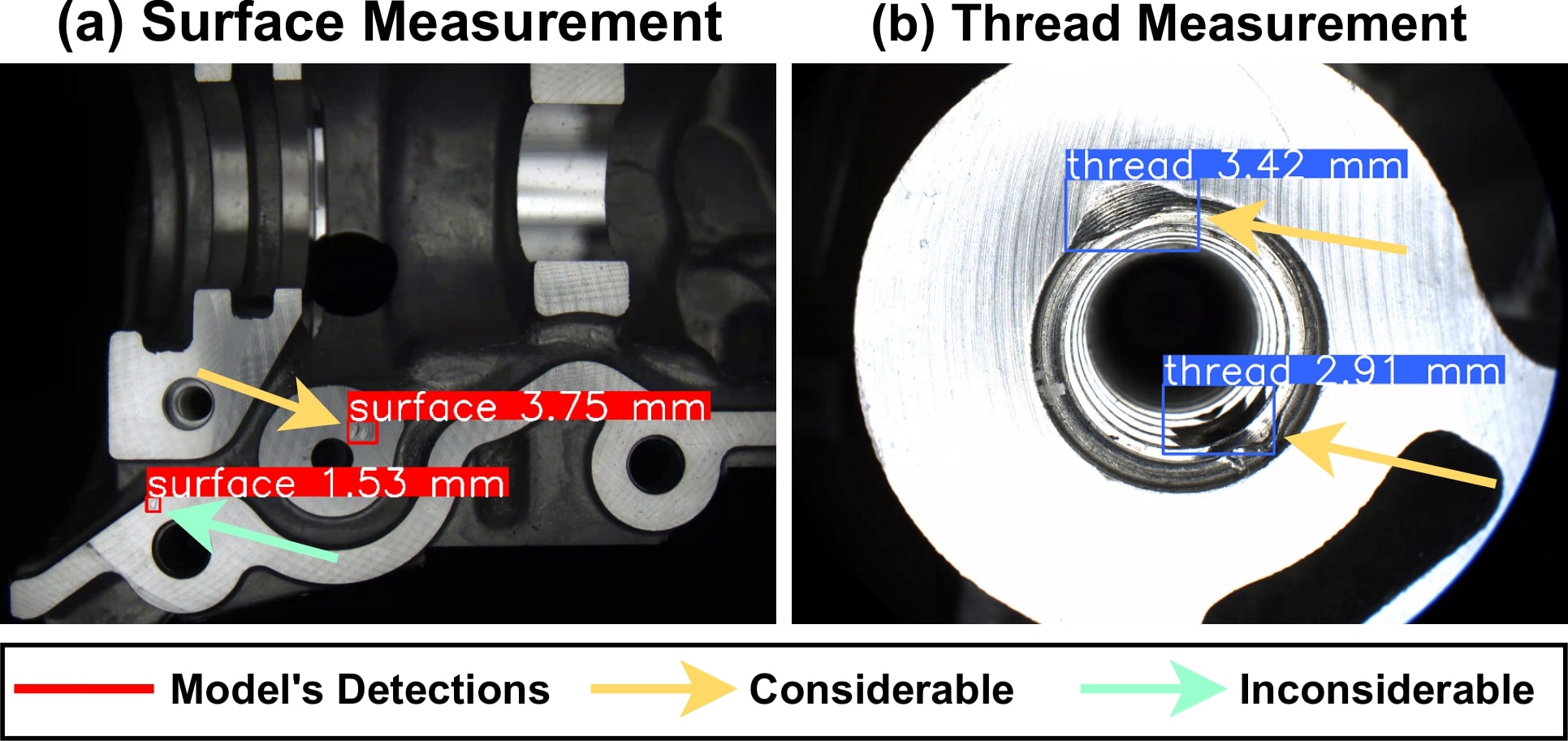} 
  \caption{Size measurement for (a) surface defects and (b) thread defects. One surface defect is below the maximum accepted size and is deemed as inconsiderable, while all the others are classified as considerable.}
  \label{fig: measurement example}
\end{figure}

The measurement module accurately estimated the dimensions of the detected defects, showing a high level of consistency with the true defect diameters, achieving a Mean Absolute Error (MAE) of $0.2$ mm. Provided that the established threshold is $2$ mm, defects measured within the $1.6$ mm to $2.4$ mm range require additional verification by a human supervisor to ensure accurate classification. Analysis of the measurement results revealed that the majority of detected defects were considerable. However, a few defects fell below the $2.0$ mm threshold, thereby proven to be inconsiderable (see Fig. \ref{fig: measurement example}). Additionally, one of the two false-positive detections measured $1.6$ mm, placing it within the acceptable size range and thus ignoring it. Consequently, only one out of the eight evaluated parts was determined to be acceptable based on the quality control standards.

\section{Conclusion}
This paper presented a novel robotic inspection solution for automated quality control in the automotive industry. By integrating a pair of collaborative robots, each equipped with a high-resolution vision system and optimized lighting, the proposed solution achieves real-time performance with high detection accuracy, ensuring reliable inspection of both surface and threads in aluminum HPDC components. Additionally, the system provides precise localization and size estimation of the detected defects. Experimental results demonstrated that the combination of advanced techniques, including image slicing, bounding box merging, and model ensemble strategies, significantly enhances detection performance, while defect size quantification further improves precision. The system is scalable and adaptable to various manufacturing applications, offering the potential to revolutionize quality control in the automotive industry. Future work will focus on developing a robotic manipulation system to enable access to all sides of the component, while optimizing processing speed through enhanced robotic collaboration.

\section*{Acknowledgment}
This work has received funding from the European Union’s Horizon 2022 research and innovation program under grant agreement N° 101120276 (SoliDAIR). A special thanks to “CIE Automotive" for their support in providing the components for this research.

\end{document}